\documentclass{article}

\usepackage{microtype}
\usepackage{graphicx}
\usepackage{subfigure}
\usepackage{booktabs} %

\usepackage{hyperref}

\usepackage{algorithm}
\usepackage{algorithmic}
\usepackage{xcolor} %
\usepackage{amsmath} %

\usepackage[accepted]{icml2025}

\usepackage{amsmath}
\usepackage{amssymb}
\usepackage{mathtools}
\usepackage{amsthm}

\usepackage{graphicx}
\usepackage{caption}
\usepackage{subcaption}
\usepackage{float}

\usepackage[capitalize,noabbrev]{cleveref}

\theoremstyle{plain}

\theoremstyle{definition}

\theoremstyle{remark}

\usepackage[textsize=tiny]{todonotes}

\definecolor{dark2green}{rgb}{0.1, 0.65, 0.3}

\usepackage{color}
\usepackage{amsmath,amssymb,amsfonts}
\usepackage{booktabs}       %
\usepackage{multirow, varwidth}
\usepackage{mathtools}
\usepackage{verbatim}
\usepackage{flushend}
\usepackage{url}
\usepackage{enumitem}
\usepackage{xcolor}         %
\usepackage{kotex}
\usepackage{colortbl}

\DeclareRobustCommand\onedot{\futurelet\@let@token\@onedot}
\def\onedot{.} %
\def\eg{\emph{e.g}\onedot, }

 \def\vs{\emph{vs}\onedot}

\newcommand{\vecx}{\mathbf{x}}

\newcommand{\mytilde}{\raise.17ex\hbox{$\scriptstyle\mathtt{\sim}$}}

\setlist[itemize]{
  itemsep=0.2em,
  parsep=0pt,
  topsep=0pt,
  partopsep=0pt
}

\icmltitlerunning{SteeringTTA: Guiding Diffusion Trajectories for Robust Test-Time-Adaptation}

\begin{document}

\twocolumn[
\icmltitle{SteeringTTA: Guiding Diffusion Trajectories \\
for Robust Test-Time-Adaptation}

\icmlsetsymbol{equal}{*}

\begin{icmlauthorlist}
\icmlauthor{Jihyun Yu}{ewha}
\icmlauthor{Yoojin Oh}{ewha}
\icmlauthor{Wonho Bae}{ubc}
\icmlauthor{Mingyu Kim}{ubc,equal}
\icmlauthor{Junhyug Noh}{ewha,equal}
\end{icmlauthorlist}

\icmlaffiliation{ewha}{Department of Artificial Intelligence, Ewha Womans University, Seoul, Republic of Korea}
\icmlaffiliation{ubc}{Department of Computer Science, University of British Columbia, Vancouver, Canada}

\icmlcorrespondingauthor{Mingyu Kim}{mgyu.kim@ubc.ca}
\icmlcorrespondingauthor{Junhyug Noh}{junhyug@ewha.ac.kr}

\icmlkeywords{Machine Learning, ICML}

\vskip 0.3in
]

\printAffiliationsAndNotice{\icmlEqualContribution}

\begin{abstract}
Test-time adaptation (TTA) aims to correct performance degradation of deep models under distribution shifts by updating models or inputs using unlabeled test data. Input-only diffusion-based TTA methods improve robustness for classification to corruptions but rely on gradient guidance, limiting exploration and generalization across distortion types. We propose \emph{SteeringTTA}, an inference-only framework that adapts Feynman-Kac steering to guide diffusion-based input adaptation for classification with rewards driven by pseudo-label. SteeringTTA maintains multiple particle trajectories, steered by a combination of cumulative top-$K$ probabilities and an entropy schedule, to balance exploration and confidence. On ImageNet‐C, SteeringTTA consistently outperforms the baseline without any model updates or source data.
\end{abstract}

\section{Introduction}
\label{sec_introduction}

\graphicspath{{src/}}   
\begin{figure*}[t]
  \centering
  \includegraphics[width=1\linewidth]{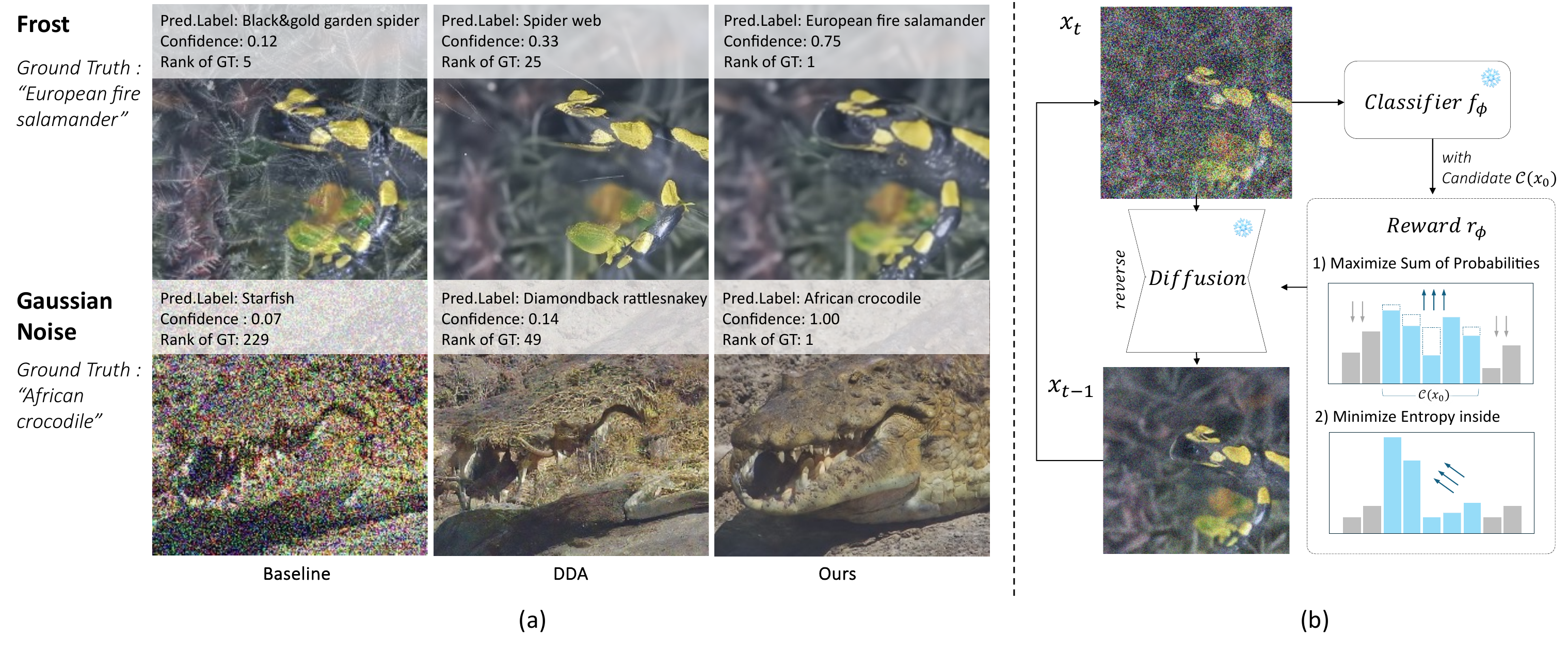} 
  \caption{
  \textbf{(a) Our motivation:} from left to right, the original corrupted image, the DDA‐denoised result, and our method's reconstruction, with corresponding ResNet‐50 predictions.  
  \textbf{(b) Pipeline overview:} at each diffusion step the image is updated, the classifier output is used to compute a reward, and the update is steered to maximize this reward.
  }
  \label{fig:motivation}
\end{figure*}

Deep networks often suffer drastic performance drops under distribution shifts at test time. 
Test‐time adaptation (TTA) methods -- either updating model weights or refining inputs on‐the‐fly -- can help, but existing diffusion‐based input adaptation methods still fail when corruptions lie in the low‐frequency band. 
For instance, Diffusion‐Driven Adaptation (DDA)~\cite{gao2023back} preserves low‐frequency structure via ILVR (Iterative Latent Variable Refinement)~\cite{choi2021ilvr}.
While effective against high‐frequency noise, it can amplify low‐frequency corruptions (\eg frost), leading to semantic distortion and inaccurate classification.

\Cref{fig:motivation} shows a frost‐corrupted image of a \textit{European fire salamander}.
A ResNet‐50 classifier (trained on clean images) classifies it as \textit{black \& gold garden spider}, with \textit{European fire salamander} in its Top‐5 predictions -- an understandable confusion.
After DDA's low‐frequency denoising, however, the amplified frost pattern leads to a wildly incorrect Top‐1 prediction \textit{spider web}, and \textit{European fire salamander} falls out of the Top‐25.
This example highlights that uniformly restoring all low‐frequency components can eliminate semantic cues.

To address this, we ask: how can we restore images without amplifying signals introduced by corruption,
while still sharpening the classifier's plausible confusion set?
We answer by \emph{steering} diffusion sampling via Sequential Monte Carlo (SMC) and Feynman-Kac potentials~\cite{singhal2025general}, guided by a pseudo‐label reward that (1) initially preserves the original confusion group and (2) gradually focuses on the correct class.

In this work, we propose \textbf{SteeringTTA}, the first diffusion‐based TTA framework to harness SMC steering.
SteeringTTA runs entirely at inference, maintains multiple particle trajectories, and uses a dynamic reward to guide resampling and denoising. 
As Figure~\ref{fig:motivation} demonstrates, our method recovers an image that the classifier correctly labels as \textit{European fire salamander}, even under severe low‐frequency corruption.

Our contributions are summarized as follows:
\begin{itemize}[leftmargin=*]
  \item We pinpoint the failure of DDA-style adaptation for low‐frequency corruption, and motivate targeted steering.  
  
  \item We introduce \emph{SteeringTTA} that leverages Feynman-Kac steering in diffusion restoration to selectively suppress corruption frequencies while preserving object semantics.

  \item We design a scheduling for pseudo‐label reward that first maintains the classifier's confusion set and then anneals entropy to converge to the true label.  
  \item We demonstrate on ImageNet‐C that SteeringTTA outperforms DDA, reliably recovering correct classes under challenging low‐frequency corruptions.
\end{itemize}

\section{Related Work}
\label{sec:related_work}

\paragraph{Test‐Time Adaptation.}
Methods either update model parameters (\eg entropy minimization~\cite{wang2020tent}) or refine inputs via self‐supervision. 
Parameter updates risk hyperparameter sensitivity and collapse, while input‐only approaches may not directly optimize accuracy.

\paragraph{Diffusion‐Based Input Adaptation.}
Diffusion purifiers restore corrupted inputs: DiffPure~\cite{nie2022diffusion} for adversarial examples, DDA~\cite{gao2023back} via low‐frequency denoising, and GDA~\cite{tsai2024gda} with gradient‐based style/semantic constraints. 
These enhance robustness but rely on differentiable guidance.

\paragraph{SMC and FK Steering.}
Sequential Monte Carlo steers diffusion by evolving multiple particles and resampling with arbitrary potentials~\cite{del2006smc}. 
Feynman–Kac steering extends this to non‐differentiable rewards without retraining~\cite{singhal2025general}.

\section{Our Approach}
\label{sec_method}

We introduce \emph{SteeringTTA}, a test‐time adaptation method that steers a pretrained diffusion model toward classification objectives using FK potentials.

After first reviewing diffusion models and SMC, we describe how FK steering is adapted for input restoration and detail our design choices -- resampling, proposal kernel, weighting -- and our pseudo‐label reward.

\subsection{Preliminaries}
\label{sec:prelim}

\paragraph{Diffusion Models.}
A diffusion model defines a forward noising process with variance $\beta_t$:
\[
q(\vecx_t\mid\vecx_{t-1}) = \mathcal{N}\bigl(\sqrt{1-\beta_t}\,\vecx_{t-1},\,\beta_t\,\mathbf{I}\bigr),\quad t=1,\dots,T,
\]
and trains a denoiser $\epsilon_\theta$ to approximate the reverse process:
\[
p_\theta(\vecx_{t-1}\mid\vecx_t).
\]
Starting from $\vecx_T\sim\mathcal{N}(0,\mathbf{I})$, repeated denoising yields a sample $\vecx_0$ close to the training distribution \cite{ho2020denoising}.

\paragraph{Sequential Monte Carlo (SMC).}
SMC approximates a sequence of target distributions $\{\pi_t\}_{t=0}^T$ by maintaining $K$ weighted particles $(\vecx_t^{(k)},w_t^{(k)})$. At each reverse step:
\begin{enumerate}[leftmargin=*]

    \item \textbf{Resampling:}
    normalize $\{w_t^{(k)}\}$, resample the particles accordingly, and begin the next proposal step with uniform weights.
 
    \item \textbf{Proposal:} sample backward each particle from defined proposal distribution $m_t$
    \[
      \vecx_{t-1}^{(k)} \sim m_t(\vecx_{t-1}\mid\vecx_t^{(k)}).
    \]
    
    \item \textbf{Weighting:} compute importance weights to correct the discrepancy between $m_t$ and $\pi_{t-1}$
    \[
      w_{t-1}^{(k)} = w_t^{(k)} \frac{\pi_{t-1}(\vecx_{t-1}^{(k)})\,m_t(\vecx_t^{(k)}\mid\vecx_{t-1}^{(k)})}{\pi_t(\vecx_t^{(k)})\,m_t(\vecx_{t-1}^{(k)}\mid\vecx_t^{(k)})}.
    \]    

\end{enumerate}
As $K\to\infty$, SMC recovers exact samples from $\pi_0$.

\paragraph{Feynman-Kac (FK) Steering.}

To bias diffusion toward high-reward outputs, define a tilted target distribution:
\begin{equation}
\label{eq:rewardtotarget}
    p_{\text{target}}(\vecx_0) \;\propto\; p_{\theta}(\vecx_0)\cdot\exp \bigl(\lambda\cdot r(\vecx_0)\bigr),
\end{equation}
with reward $r(\vecx_0)$ and scale $\lambda>0$. FK steering introduces potentials $G_t$ which tilt the distribution as
\begin{equation}
\label{eq:potentialqual}
    p_{\mathrm{FK}}(\vecx_{0:T}) \;\propto\; p_{\theta}(\vecx_{0:T}) \prod_{t=0}^{T-1} G_t\bigl(\vecx_{t:T}\bigr),
\end{equation}

ensuring the marginal of $\vecx_0$ follows $p_{\mathrm{target}}$ by setting $\prod_{t=0}^{T-1}G_t(\vecx_{t:T})=\exp\bigl(\lambda\cdot r(\vecx_0)\bigr)$.
In practice, we estimate $\vecx_0$ with intermediate states of reverse diffusion processes via Tweedie’s formula~\cite{chung2022diffusion, efron2011tweedie}:
\begin{equation}
\hat\vecx_t = \frac{\vecx_t - \sqrt{1-\alpha_t}\,\epsilon_\theta(\vecx_t)}{\sqrt{\alpha_t}},
\end{equation}
where $\alpha_t=\prod_{s=1}^{t}(1-\beta_s)$, and evaluate $r(\hat\vecx_t)$ to compute $G_t$. 
No backpropagation through $r$ is required, allowing non-differentiable and classification-aware rewards to steer sampling processes.

\subsection{Generalized FK steering for TTA}
\label{sec:fk}

Given a corrupted input $\vecx_0$, we first perform $N$ forward noising steps to reach $\vecx_N$.
We then run $N$ reverse steps, tracking $K$ parallel particles. At each timestep $t$:

\begin{enumerate}[leftmargin=*]

\item \textbf{Resampling.}

Following \citet{singhal2025general}, we monitor particle degeneracy via the \emph{effective sample size}
\(\mathrm{ESS}_t = \bigl[\sum_{i=1}^K (\hat G_t^i)^2\bigr]^{-1}\),
where \(\hat G_t^i\) are the normalized FK potentials. 
If \(\mathrm{ESS}_t < K/2\), we trigger multinomial resampling to preserve diversity without extra computation.

\item \textbf{Proposal.}
FK-Steering is compatible with any proposal kernel $\tau(\vecx_{t-1}\!\mid\!\vecx_t)$, \eg the unbiased reverse kernel or gradient-guided variants. Unlike typical generative tasks where we generate i.i.d samples of the data distribution, we need to preserve the semantics of the given image for the following classification. To this end, we employ a low-pass filter that maintains the overall semantics of an image~\citep{gao2023back,raman2023turn}.

\item \textbf{Weighting.}
Among potentials that satisfy
Eq.~\eqref{eq:potentialqual}, we employ the \textit{difference} potential following \cite{singhal2025general, wu2023practical},
\begin{equation}
\label{eq:potentialdiff}
    G_t(\vecx_{t-1},\vecx_{t}) = \text{exp}(\lambda \cdot (r_\phi(\hat\vecx_{t-1})-r_\phi(\hat\vecx_{t}))) 
\end{equation}
which directly rewards particles whose predicted clean image
$\hat{\vecx}_{t-1}$ improves the classifier’s objective.

\end{enumerate}

We maintain $K$ particles, computing rewards on intermediate $\hat 
\vecx_t^{(k)}$, and iteratively steering and pruning them.
At $t=0$, the particle with the highest reward is chosen as the adapted image;
please refer to \cref{alg:fkd-steering} for details.

\hbadness=99999
\begin{algorithm}[tb]
\caption{SteeringTTA}
\label{alg:fkd-steering}
\begin{algorithmic}[1]
\STATE \textbf{Input:} Corrupted image $\vecx_0$, Diffusion model $p_\theta({\vecx}_{0:T})$, Classifier $f_\phi$, Potentials $G_t$, Proposal $\tau(\vecx_t \mid \vecx_{t+1})$, Reward function $r_\phi(\cdot)$, Number of particles $K$

\STATE $N$: diffusion range 
\STATE $\phi_D$: low-pass filter of scale D
\STATE $\hat\vecx_t$: predicted clean image at timestep t
\STATE $\mathcal C$ : set built with $f_\phi(\vecx_0)$.
\STATE Sample $\vecx_N^i \sim q(\vecx_N \mid \vecx_0)$ \hfill 
\textcolor{blue!80}{\(\triangleright\) forward pass}
\STATE Define $r_\phi(\cdot) \gets r_\phi(\cdot, t, \mathcal C)$ \hfill\textcolor{blue!80}{\(\triangleright\) Eq.\,\ref{eq:final_reward}}
\STATE Initial weights, $G_N^i = \exp(\lambda \cdot 
r_\phi(\hat\vecx_N^i))$ for $i \in [K]$
\FOR{$t = N-1$ {\bfseries to} $0$} 
    \STATE \textbf{Resample:} 
    \STATE Sample indices $a_t^i \sim \text{Multinomial}({\vecx}_t^i, G_t^i)$ 
    \STATE and set ${\vecx}_t^i = {\vecx}_t^{a_t^i}$ for $i \in [K]$
    \STATE \textbf{Propose:} \hfill \textcolor{blue!80}{\(\triangleright\) low-pass filtering}    
    \STATE $\vecx_{t-1}^i, \hat{{\vecx}}_{t-1}^i \sim \tau_\theta({\vecx}_{t-1} \mid {\vecx}_t)$ 
    \STATE ${\vecx}_{t-1}^i \gets \vecx_{t-1}^i - {w} \nabla_{{\vecx}_t} \left\| \phi_D({\vecx}_0) - \phi_D(\hat{{\vecx}}_{t-1}^i) \right\|_2$
    
    \STATE \textbf{Weight:}  
    \STATE $G_{t-1}^i$ for $i \in [K]$ using:
    \[
    G_{t-1}^i = \frac{p_\theta({\vecx}_{t-1}^i \mid {\vecx}_t^i)}{\tau({\vecx}_{t-1}^i \mid {\vecx}_t^i)} 
    G_{t-1}({\vecx}_N^i, \ldots, {\vecx}_{t-1}^i)
    \]
\ENDFOR
\STATE \textbf{Output:} $\displaystyle \vecx_0^g \gets \arg\max_{i \in [K]} r_\phi(\hat\vecx_0^i)$
\end{algorithmic}
\end{algorithm}

\subsection{Test-time Reward}
\label{sec:reward}

\paragraph{Candidate set construction.}  
As ground-truth labels are unavailable at test time, we define an adaptive candidate set,
\begin{equation}
\label{eq:candidate}
  \mathcal{C}(\vecx_0)
  = \bigl\{\,y : \sum_{y'\in\mathrm{desc}(y)} p(y' \mid \vecx_0) \ge P\%\bigr\},
\end{equation}
accumulating classes in descending order of predicted probability on the corrupted input until their total mass exceeds $P\%$.
This preserves the classifier's initial confusion group.

\paragraph{Reward formulation.}
Our steering reward at step \(t\) is
\begin{equation}
\label{eq:final_reward}
\resizebox{0.90\columnwidth}{!}{$
    \begin{aligned}
        r_\phi(\hat\vecx_t, t, \mathcal{C})= \left( 1 - \alpha(t) \right) \log\sum_{y \in \mathcal{C}(\vecx_0)} p(y \mid \hat\vecx_t) - \alpha(t)H\bigl(\mathcal{C}(\vecx_0)\bigr),
    \end{aligned}
    $}
\end{equation}

with an entropy on the classes in the adaptive candidate set,
\begin{equation}
\label{eq:entropy}
    \begin{aligned}
        H\bigl(\mathcal{C}(\vecx_0)\bigr) = -\sum_{y \in \mathcal{C}(\vecx_0)} p(y \mid \hat\vecx_t) \log p(y \mid \hat\vecx_t).
    \end{aligned}
\vspace{-0.07in}
\end{equation}

The \emph{log‐sum} term encourages boosting total probability over the plausible labels by steering it to be close to $1$, preventing jumps to unrelated classes; the \emph{entropy} term discourages the predicted probability to be an uniform distribution, pushing the sampler to commit to a single label.

\paragraph{Annealing schedule.}
We linearly anneal $\alpha(t)$ from $0$ at $t=N$ (favoring exploration via log‐sum) to $1$ at $t=0$ (favoring exploitation via entropy minimization).
Early iterations thus maintain the original confusion set, while later ones refine focus on the correct class.

\begin{table*}[ht!]
\caption{
Average top‐1 accuracy (\%) on severity $5$ ImageNet‐C.  
``Baseline'' denotes corrupted images evaluated without any adaptation; other methods are described in the main text. 
See Appendix~\ref{sec:supp:abbr} for corruption abbreviation definitions.
}
\label{tab:overall_results}
\centering
\footnotesize
\resizebox{\textwidth}{!}{%
\begin{tabular}{l*{16}{r}}   
\toprule
\multirow{2}{*}{\textbf{Method}} & 
\multicolumn{4}{c}{\textbf{Blur}} &
\multicolumn{4}{c}{\textbf{Digital}} &
\multicolumn{3}{c}{\textbf{Noise}} &
\multicolumn{4}{c}{\textbf{Weather}} &
\multirow{2}{*}{\textbf{Avg.}} \\
\cmidrule(lr){2-5}\cmidrule(lr){6-9}\cmidrule(lr){10-12}\cmidrule(lr){13-16}
 & Def.\ & Glass & Mot.\ & Zoom
 & Contr.\ & Elast.\ & JPEG & Pixel
 & Gauss.\ & Impl.\ & Shot
 & Bright & Fog & Frost & Snow & \\
\midrule
Baseline \scriptsize{(w.o. Adaptation)} 
& 12.00 &  5.80 & 12.80 & 21.20 &  \textbf{3.00} & 14.60 & 43.20 & 30.40 &  5.20 &  6.00 &  8.20 & 54.80 & \textbf{21.40} & 20.80 & 16.80 & 18.41  \\

Diffpure \scriptsize~\cite{nie2022diffusion}
& 2.00 &  6.20 & 5.60 & 9.80 &  0.40 & 18.60 & 43.20 & 24.60 &  5.00 &  4.40 &  6.00 & 42.60 & 1.60 & 12.80 & 10.00 & 12.85  \\

DDA\scriptsize~\cite{gao2023back}
& \textbf{12.40} & 10.20 & 13.00 & 23.40 &  2.80 & \textbf{34.20} & 50.20 & 47.80 & 50.40 & 49.60 & 51.40 & \textbf{55.00} & 18.60 & 27.60 & 19.40 & 31.07 \\

Grad-DDA  
& \textbf{12.40} & \textbf{11.00} & \textbf{13.20} & \textbf{24.40} &  2.40 & 32.20 & \textbf{50.80} & \textbf{51.40} & 49.00 & 49.20 & 50.20 & 53.60 & 16.40 & 28.20 & 20.60 & 31.00 \\

SteeringTTA (Ours) 
& 12.20 & \textbf{11.00} & 12.80 & 23.80 &  2.40 & 33.80 & 49.00 & 49.80 & \textbf{52.60} & \textbf{50.60} & \textbf{53.40} & 52.80 & 18.40 & \textbf{29.40} & \textbf{21.60} & \textbf{31.57} \\

\rowcolor{gray!15}
SteeringTTA (GT) 
& 12.40 & 10.60 & 13.40 & 26.00 &  2.20 & 34.60 & 50.20 & 51.80 & 57.40 & 57.40 & 58.60 & 53.80 & 18.20 & 30.80 & 24.40 & 33.45 \\
\bottomrule
\end{tabular}}
\end{table*}

\section{Experiments}
\label{sec_experiment}

\subsection{Experimental Settings}

\paragraph{Dataset.}
We evaluate on ImageNet‐C~\cite{hendrycks2019robustness}, which applies fifteen corruption types at five severity levels to the ImageNet validation set. 
Following the small‐subset protocol, we randomly sample one image per class for $100$ classes, forming five disjoint splits ($1{,}500$ images per split at severity $5$).
We report mean of Top‐1 accuracy across these splits.

\paragraph{Models.}
For input restoration, we use a pretrained unconditional diffusion model ($256\times256$) trained on ImageNet-1K~\cite{dhariwal2021diffusion}.
The classifier is fixed with the ResNet-50~\cite{he2016deep} trained on ImageNet.
During TTA, the classifier provides pseudo-labels for steering and serves as the evaluator -- its weights remain frozen, preventing information leakage from the target data.

\paragraph{Baselines.}
We compare SteeringTTA against:
\begin{itemize}[leftmargin=*]
  \item \textbf{Diffpure}~\cite{nie2022diffusion}: Diffusion based adversarial purification that simply adds a small amount of noise and then reverses until $t=0$.
  \item \textbf{DDA}~\cite{gao2023back}: Diffusion‐Driven Adaptation with default hyperparameters.
  \item \textbf{Gradient-guided DDA (Grad-DDA)}: an ablation that steers DDA's reverse process via our pseudo-label reward using gradient ascent.
  \item \textbf{SteeringTTA (GT)}: SteeringTTA using the ground-truth log-likelihood $\log p(y|x)$ as the reward; upper-bound of improvement from SteeringTTA.
\end{itemize}

\paragraph{Implementation Details.}
We use \(K=4\) particles and \(N=50\) reverse steps, and follow the resampling schedule of \citet{singhal2025general}: resample whenever \(\mathrm{ESS}<K/2\) at every $5$ steps.  
Our test‐time reward combines the cumulative Top‐\(K\) probability using adaptive threshold \(P=70\%\), with an entropy term, and employs a linear annealing schedule \(\alpha(t)\!:0\!\to\!1\).  
We set the reward scale \(\lambda=1\).
We use an ensemble for the final predictions~\citep{gao2023back}, taking $\arg\!\max_{y}\; \tfrac12\!\bigl(p(y|\vecx_0) + p(y|\vecx_0^{\,g})\bigr)$, the most probable class predicted from the original \(\vecx_0\) and adapted image \(\vecx_0^{\,g}\).
All hyperparameters were fixed a priori (no per‐image tuning); each experiment was run once on a NVIDIA A100 GPU.

\subsection{Experimental Results}

\paragraph{Overall Performance.}
Table~\ref{tab:overall_results} presents average Top-1 accuracy on severity $5$ ImageNet-C. 
SteeringTTA outperforms DDA by \(0.51\%\), while Grad-DDA shows no significant gain. 

We hypothesize that this is due to Grad-DDA’s higher sensitivity to noise in the guidance.
Although both methods use the same guidance, Grad-DDA applies it by injecting gradients directly into the pixel space along a single trajectory, which can amplify label noise into high-frequency artifacts {\ref{fig:app_figure_grad}}.
In contrast, SteeringTTA utilizes guidance through resampling and weighting, avoiding direct pixel-level updates.

GT-guided SteeringTTA further improves by \(2.39\%\), indicating the gap to the ideal upper-bound. 
These results confirm that FK steering with pseudo-label rewards is more effective than gradient guidance alone for robust TTA.

\paragraph{Category‐wise gains.}
Table~\ref{tab:cat_gain_small} reports relative improvements over DDA by corruption type.
Gradient‐guided adaptation shows mixed results; it improves blur and digital but degrades noise and weather.
GT-guided steering delivers large gains (up to $+7.3\%$), particularly for noise.
Our SteeringTTA yields consistent, moderate improvements except for digital; $0.51\%$ gain on average.

\begin{table}[!t]
\caption{Performance gain \vs\ DDA per corruption category.}
\label{tab:cat_gain_small}
\centering
\footnotesize
\setlength{\tabcolsep}{9pt}
\begin{tabular}{cccc>{\columncolor{gray!15}}c}
\toprule
\textbf{Category} &
\textbf{DDA} &
\textbf{Grad-DDA} &
\textbf{Ours} &
\textbf{GT} 
\\
\midrule
Blur     
& 14.75 & \textcolor{blue}{$+0.50$} &  \textcolor{blue}{$+0.20$}&  \textcolor{blue}{$+0.85$}  \\
Digital  
& 33.75 & \textcolor{blue}{$+0.45$} &  \textcolor{blue}{$+0.00$}&  \textcolor{blue}{$+0.95$}  \\
Noise    
& 50.47 & \textcolor{red}{$-1.00$} &   \textcolor{blue}{$+1.73$} &\textcolor{blue}{$+7.33$}  \\
Weather  
& 30.15 & \textcolor{red}{$-0.45$} &  \textcolor{blue}{$+0.40$} & \textcolor{blue}{$+1.65$}  \\
\midrule
\textbf{Average} & 31.07 & \textcolor{red}{$-0.07$} &  \textcolor{blue}{$+0.51$}&  \textcolor{blue}{$+2.39$}  \\
\bottomrule
\end{tabular}
\end{table}

\section{Conclusion}
\label{sec_conclusion}

We present \emph{SteeringTTA}, a novel test‐time adaptation framework that steers a pretrained diffusion model using Feynman-Kac potentials.
By integrating a pseudo‐label‐driven reward into the reverse diffusion process, SteeringTTA directly optimizes classification accuracy -- unlike prior methods that rely on surrogate objectives or gradient‐only guidance.
Our multi‐particle sampling explores diverse hypotheses and selects high‐reward paths to prevent collapse.
On ImageNet‐C, SteeringTTA outperforms DDA by $0.51\%$ top‐1 accuracy, validating the effectiveness of reward‐based steering. 
Future work includes adaptive resampling, richer reward designs, and applications to other domains and real‐world shifts.  

\nocite{langley00}

\bibliography{icml2025}
\bibliographystyle{icml2025}

\clearpage
\appendix

\setcounter{equation}{0}
\renewcommand{\theequation}{\Alph{section}.\arabic{equation}}
\setcounter{table}{0}
\renewcommand{\thetable}{\Alph{section}.\arabic{table}}
\setcounter{figure}{0}
\renewcommand{\thefigure}{\Alph{section}.\arabic{figure}}

\section{Corruption Abbreviations}
\label{sec:supp:abbr}

This section provides the full names of the corruption-type abbreviations used in Table~\ref{tab:overall_results}, to help interpret the per-category results.

\begin{table}[!h]
\centering
\small
\setlength{\tabcolsep}{10pt}
\caption{Definitions of corruption abbreviations.}
\begin{tabular}{ll}
\toprule
\textbf{Abbrev.} & \textbf{Corruption}\\
\midrule
Def.    & Defocus blur \\
Glass   & Glass blur \\
Mot.    & Motion blur \\
Zoom    & Zoom blur \\
Contr.  & Contrast \\
Elast.  & Elastic transform \\
JPEG    & JPEG compression \\
Pixel   & Pixelate \\
Gauss.  & Gaussian noise \\
Impl.   & Impulse noise \\
Shot    & Shot noise \\
Bright  & Brightness change \\
Fog     & Fog \\
Frost   & Frost \\
Snow    & Snow \\
\bottomrule
\end{tabular}
\end{table}

\section{Ablation Studies}

\subsection{Effects of hyperparamters} 

\paragraph{Reward Coefficient $\lambda$.} 
The coefficient \(\lambda\) rescales the reward in Eq.~(\ref{eq:rewardtotarget}) which is then exponentiated by the difference-potential of Eq.~(\ref{eq:potentialdiff}) to obtain particle weights.  
With the candidate–set threshold fixed at \(P=70\%\), the reward values approximately lie in \([-0.25,\,0]\); the corresponding potential is therefore upper-bounded by \(0.25\), so larger \(\lambda\) values exponentially attenuate the weights of low-reward particles at very early steps.  
Table~\ref{tab:abla_lambda_p} lists Top-1 accuracy on ImageNet-C for several \(\lambda\) values, confirming that the default setting \(\lambda=1\) consistently outperforms \(\lambda=5\) across all variants on average.

\paragraph{Adaptive Threshold $P$.}
Table \ref{tab:abla_lambda_p} contrasts $P\!\in\!\{50\%,70\%\}$ showing $P=70$ consistently outperforms which means that in $P=50$, the ground-truth (GT) label is frequently excluded -- expected because corrupted inputs place the GT label deep in the posterior tail.  

\begin{table}[h]
\centering
\small
\setlength{\tabcolsep}{7pt}
\caption{Ablation on varying $\lambda$ and adaptive threshold $P$.}
\label{tab:abla_lambda_p}
\begin{tabular}{ccccccc}
\toprule
\multicolumn{2}{c}{\textbf{Settings}} & \multicolumn{4}{c}{\textbf{Category}} & \multirow{2}{*}{\textbf{Avg.}} \\
\cmidrule(lr){1-2}\cmidrule(lr){3-6}
\(\lambda\) & \(P\) & Blur & Digital & Noise & Weather &  \\
\midrule
1 & 50\% & \textbf{15.30} & 33.20 & 52.07 & 29.95 & 31.33 \\
5 & 50\% & 14.80 & \textbf{33.95} & 50.93 & 29.05 & 30.93 \\
1 & 70\% & 14.95 & 33.75 & \textbf{52.20} & \textbf{30.55} & \textbf{31.57} \\
5 & 70\% & 15.15 & 32.45 & 50.07 & 29.45 & 30.56 \\
\bottomrule
\end{tabular}
\end{table}

\subsection{Robustness Across ImageNet-C Subsets} 
We randomly sample the ImageNet-C into five disjoint splits, each comprising $100$ classes across all $15$ corruption types.
As Table~\ref{tab:subset_results} shows, our method attains the best Top-1 accuracy on every split, demonstrating that its superiority is robust and not merely the result of a fortuitous partition.

\begin{table}[h]
\caption{Comparison of results on $5$ ImageNet-C subsets.}
\vspace{-2mm}
\label{tab:subset_results}
\centering
\small
\setlength{\tabcolsep}{5pt}
\begin{tabular}{lccccc c}
\toprule
\multirow{2}{*}{\textbf{Method}} & \multicolumn{5}{c}{\textbf{Subset}} & \multirow{2}{*}{\textbf{Avg.}} \\
\cmidrule(lr){2-6}
& 1     & 2     & 3     & 4     & 5     &            \\
\midrule
Baseline & 18.60 & 15.67 & 19.47 & 20.47 & 17.87 & 18.41      \\
Diffpure & 13.87 & 10.60 & 13.40 & 14.40 & 12.00 & 12.85      \\
DDA      & 32.27 & 28.33 & 32.27 & 33.27 & 29.20 & 31.07      \\
Grad‐DDA & 32.47 & 27.53 & 32.20 & 32.93 & 29.87 & 31.00      \\
Ours     & \textbf{33.20} & \textbf{28.80} & \textbf{32.33} & \textbf{33.47} & \textbf{30.07} & \textbf{31.57} \\
\rowcolor{gray!15}
GT       & 35.20 & 30.40 & 34.33 & 35.20 & 32.13 & 33.45      \\
\bottomrule
\end{tabular}
\end{table}

\subsection{Benefit of Post-ensemble Aggregation}

\Cref{tab:ensemble_ablation} compares the effect of the ensemble strategy for DDA, Grad-DDA, SteeringTTA (Ours) and SteeringTTA (GT).
The ensemble strategy improves the performance by about $1\text{--}2\%$ on average.
The improvement with the ensemble for different corruptions is generally not significant except for weather corruption; the increase for weather corruption is up to $+8.8\%$.

\newcolumntype{F}{>{\centering\arraybackslash}p{2.45em}}
\begin{table}[h]
\centering
\small
\fontsize{8}{10}\selectfont
\setlength{\tabcolsep}{7pt}
\caption{Effect of ensemble for the final predictions.}
\label{tab:ensemble_ablation}
\vspace{-2mm}
\begin{tabular}{lFFFFF}
\toprule
\textbf{Method} & \textbf{Blur} & \textbf{Digital} & \textbf{Noise} & \textbf{Weather} & \textbf{Avg.} \\
\midrule
DDA                         & 14.05 & 33.10 & 51.80 & 22.65 & 28.97 \\
\rowcolor{gray!15}
$+$ Ensemble              & 14.75 & 33.75 & 50.47 & 30.15 & 31.07 \\
Grad-DDA                    & 14.50 & 35.10 & 50.80 & 23.65 & 29.69 \\
\rowcolor{gray!15}
$+$ Ensemble         & 15.25 & 34.20 & 49.47 & 29.70 & 31.00 \\
Ours          & 14.25 & 33.85 & 52.87 & 21.75 & 29.20 \\
\rowcolor{gray!15}
$+$ Ensemble   & 14.95 & 33.75 & 52.20 & 30.55 & 31.57 \\
GT            & 14.80 & 35.80 & 59.20 & 26.05 & 32.28 \\
\rowcolor{gray!15}
$+$ Ensemble     & 15.60 & 34.70 & 57.80 & 31.80 & 33.45 \\
\bottomrule
\end{tabular}
\end{table}

\subsection{Impact of Classifier Guidance Scale for Grad-DDA} 

\Cref{tab:grad_dda_scale} provides an ablation study with varying guidance scale $s = \{ 1, 10 \}$ and threshold $P=\{ 50, 70 \}$.
The results show that $s=1$ is significantly better than $s=10$.
However, the gap between $P=50$ and $P=70$ is marginal in general.

\begin{table}[h]
\centering
\small
\setlength{\tabcolsep}{7pt}
\caption{Ablation on classifier scale \(s\) and threshold \(P\) for Grad‐DDA.}
\vspace{-1mm}
\label{tab:grad_dda_scale}
\begin{tabular}{ccrrrrr}
\toprule
\multicolumn{2}{c}{\textbf{Settings}} & \multicolumn{4}{c}{\textbf{Category}} & \multirow{2}{*}{\textbf{Avg.}} \\
\cmidrule(lr){1-2}\cmidrule(lr){3-6}
\(s\) & \(P\) & Blur & Digital & Noise & Weather &  \\
\midrule
1  & 50\% & \textbf{15.25} & \textbf{34.40} & 48.33 & \textbf{30.40} & \textbf{31.01} \\
10 & 50\% &  9.55           & 22.45           & 10.93 & 23.00           & 16.85           \\
1  & 70\% & \textbf{15.25} & 34.20           & \textbf{49.47} & 29.70           & 31.00           \\
10 & 70\% &  9.50           & 24.35           & 13.60 & 23.05           & 17.89           \\
\bottomrule
\end{tabular}
\end{table}

\subsection{Computational Costs} 
We compare per-image latency at the \texttt{p\_sample\_loop} kernel to isolate adaptation cost. On a single NVIDIA A100 and the brightness corruption, Grad-DDA takes  $8.0$ seconds, whereas SteeringTTA takes $13.2$ seconds. This gap stems from SteeringTTA's multiple particles: it simulates $K$ particles per image, which multiplies neural nets function evaluations (NFEs) for the backbone UNet and requires doubled overall runtime than the single-particle Grad-DDA in the case of $K=4$.

\section{Detailed Related Works}
\label{appendix:related_work}

\subsection{Diffusion Models for Test-Time Adaptation}
Test-time adaptation (TTA) using diffusion models has emerged as a promising research direction to improve robustness of a discriminative model \eg image classifier, under distribution shifts.
Recent works can be categorized into two branches based on what is adapted at test time: (1) joint adaptation that adapts both inputs and model parameters using diffusion-based feedback, and (2) input-only adaptation that refines test inputs via a diffusion model keeping the discriminative model fixed.

\paragraph{Joint Adaptation.} 

The most common way to improve robustness at test time is to update both the input and model weights simultaneously. 
\citet{raman2023turn} applies pseudo-label ensembling to refine the classifier as it transforms each test image~\citep{raman2023turn}. 
Similarly, Diffusion-TTA~\cite{prabhudesai2023diffusion} ties the classifier and diffusion model in a feedback loop: the classifier conditions the reverse diffusion process, while diffusion outputs guide small weight updates.
SDA~\cite{guo2024sda} uses a diffusion model to translate target images into a synthetic domain that mimics the source, then fine-tunes the classifier on this synthetic data so the model itself adapts.

These joint strategies may yield higher performance gains than input-only approaches, allowing a discriminative model to actively adapt to generated samples. 
However, they may be more sensitive to hyperparameters and introduce additional computational overhead.

\paragraph{Input-only Adaptation.}
Diffpure~\citep{nie2022diffusion} purifies adversarial attacks by using a small diffusion timestep. Diffpure suggests that only adding a small amount of noise and solving the reverse stochastic differential equation in diffusion could effectively wash out adversarial perturbation. However, Diffpure showed performance degradation when it applied to test time adaptation ~\citep{tsai2024gda, gao2023back}.

DDA~\citep{gao2023back} adapts test images only via a diffusion model trained on a source domain. 
It aligns corrupted images to the source domain by denoising them with low-pass filtering.

However, as DDA preserves low-frequency information only using ILVR~\citep{choi2021ilvr}, without considering following classification tasks, it often fails to recover images for correct classification.

GDA~\citep{tsai2024gda} also attempts to denoise corrupted test images using a diffusion model, but it incorporates additional style and semantic constraints in the reverse process. 

In addition, it minimizes marginal entropy during the reverse process considering downstream classification tasks.
However, as it is computed only on top-1 pseudo-labels, it may not be reliable when class predictions are wrong. 
Both DDA and GDA rely on gradient-based diffusion process, which limits the guidance to differentiable objectives.

\subsection{Sequential Monte Carlo for Diffusion Models}
 
Some recent works in diffusion models employ Sequential Monte Carlo (SMC) for more flexible sampling in the reverse process. 
In contrast to gradient-based guidance, with SMC, particles (or samples) evolve through diffusion processes; they are reweighted and resampled according to an user-defined criterion.

A practical trigger for resampling is the \emph{effective sample size} (ESS):
\[
  \mathrm{ESS}_t = \Bigl[\sum_{i=1}^{K} (\hat G_t^i)^{2}\Bigr]^{-1},
\]
where \(\hat G_t^i\) are the normalized potentials. 
When \(\mathrm{ESS}_t < 0.5K\), resampling prevents weight collapse and maintains particle diversity~\cite{singhal2025general,wu2023practical}.

By biasing the sampling distribution toward higher-potential regions, SMC can incorporate non-differentiable rewards without retraining or backpropagating through the diffusion model. 

This property is particularly appealing for test-time adaptation, where reward functions may be implicit or non-differentiable.

\citet{wu2023practical} introduces a twisted diffusion sampler offering asymptotically exact conditional generation via SMC, outperforming naive conditional heuristics on tasks like image inpainting.
\citet{kim2025test} proposes test-time Diffusion Alignment as Sampling(DAS),

which uses SMC to maximize a reward that reflects alignment to a given goal, avoiding reward over-optimization issues common in RL fine-tuning and reward under-optimization issues in gradient-guidance. 
\citet{singhal2025general} presents a comprehensive Feynman-Kac (FK) steering framework that formalizes diffusion models with SMC, allowing arbitrary, possibly non-differentiable rewards to modify the generative trajectory without updating model parameters.
Their method shows strong performance on text-to-image tasks, often rivaling specialized fine-tuned models, all via particle-based sampling. Various works also explore SMC to address domain gaps in designing biological sequence~\citep{li2024derivative}, text generation~\citep{singhal2025general} and inverse problems~\citep{cardoso2023monte}
highlighting the versatility of particle filtering strategies for diffusion models.

\onecolumn

\section{Qualitative Results}

\graphicspath{{src/}}   
\begin{figure}[h!]
  \centering
  \includegraphics[width=0.87\textwidth]{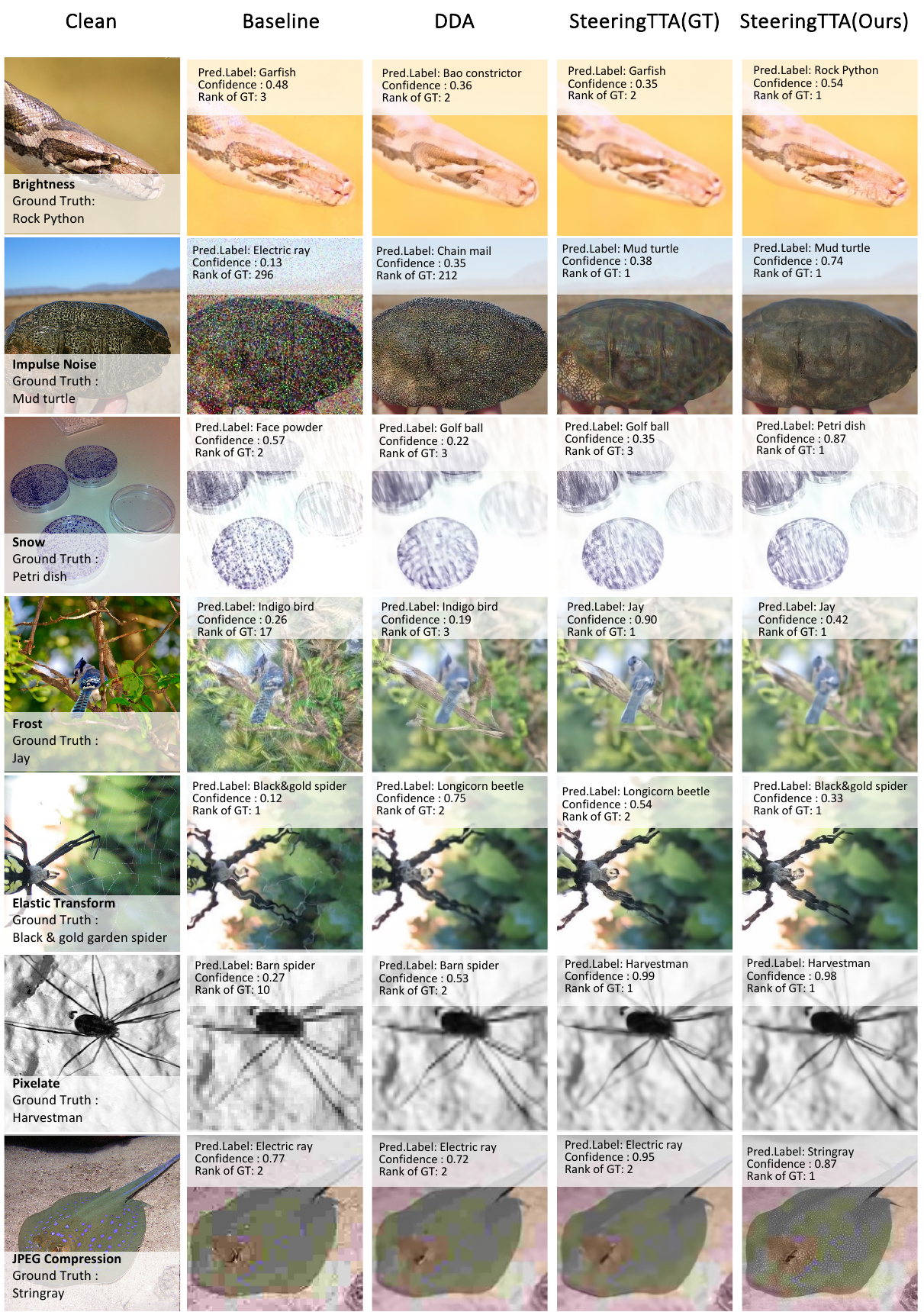} 
  \caption{Qualitative results comparing the original corrupted image, DDA, GT-based SteeringTTA and ours.  
  }
  \label{fig:app_qual}
\end{figure}

\graphicspath{{src/}}   
\begin{figure}[t!]
  \centering
  \includegraphics[width=0.55\textwidth]{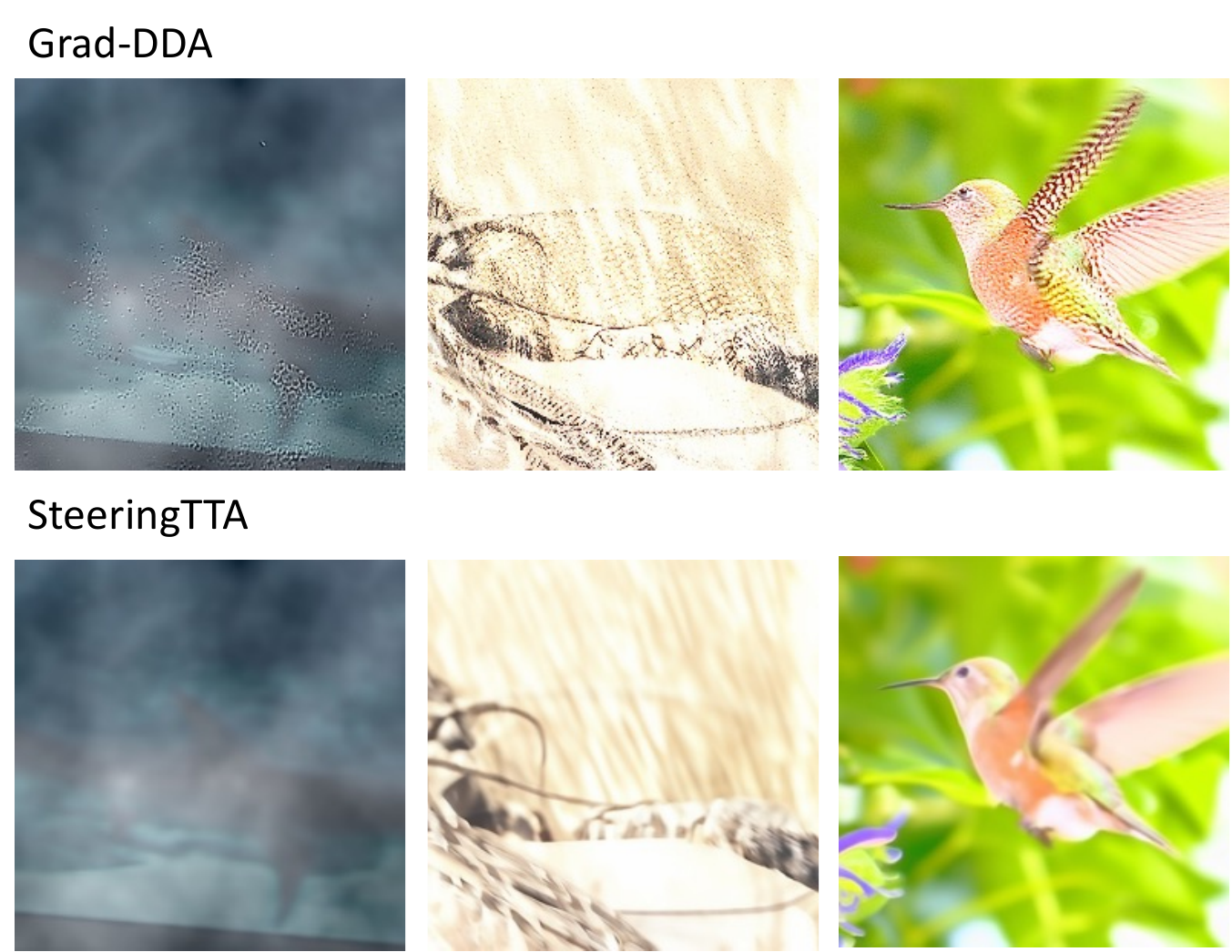} 
  \caption{From top to bottom, the adapted image with Grad-DDA ($scale=10$) and our method's with same rewards. Some unreliable high-frequency artifacts appear in Grad-DDA which are not in ours. 
  }
  \label{fig:app_figure_grad}
\end{figure}

\end{document}